\documentclass[conference]{IEEEtran}
\usepackage{amsmath,amssymb}
\usepackage{txfonts}
\usepackage{booktabs,multirow,tabularx}
\usepackage{cite}
\usepackage{graphicx}
\usepackage{url}
\usepackage[colorlinks=true,linkcolor=black,citecolor=black,urlcolor=blue,hypertexnames=false]{hyperref}
\usepackage{xcolor}
\usepackage{balance}
\usepackage{stfloats}
\usepackage{placeins}
\usepackage{float}
\usepackage{capt-of}
\graphicspath{{figures/}}
\makeatletter
\def\section{\@startsection{section}{1}{\z@}{1.5ex plus 1.5ex minus 0.5ex}%
{0.7ex plus 1ex minus 0ex}{\normalfont\normalsize\centering\scshape\bfseries}}
\def\subsection{\@startsection{subsection}{2}{\z@}{1.5ex plus 1.5ex minus 0.5ex}%
{0.7ex plus .5ex minus 0ex}{\normalfont\normalsize\bfseries\itshape}}
\def\subsubsection{\@startsection{subsubsection}{3}{\parindent}{0ex plus 0.1ex minus 0.1ex}%
{0ex}{\normalfont\normalsize\bfseries\itshape}}
\makeatother

\title{Morphology-Aware Human Motion Retargeting for Wheeled-Humanoid Loco-Manipulation}

\author{\IEEEauthorblockN{Chenbo Xia, Xiang Li, Hao Zhang, Bohan Wu, Yue Hao, and Chao Ye\textsuperscript{*}}
\IEEEauthorblockA{Harbin Institute of Technology}}

\begin{document}
\maketitle

\begin{abstract}
We propose a whole-body teleoperation method for topology-mismatched humanoids, with a focus on motion retargeting and whole-body control for wheeled humanoids. First, we design a Topology-Adaptive Motion Retargeting (TAMR) method with a robust switching logic that selects mapping joints according to human posture. Second, we train a Teacher--Student DAgger method that supports real-time teleoperation with inertial motion capture and virtual-reality devices. We further develop a Temporal Feasibility-Aware Command Governor (T-FeaCG) for the wheeled base to reduce the mismatch between teleoperation policy intent and the feasible execution margin of a three-steer-wheel base. We will release the complete teleoperation code and dataset, and demonstrate whole-body humanoid skills learned from teleoperation. More videos are available at: \protect\href{https://xiachenbuo.github.io/Morphology-Aware-Human-Motion-Retargeting-for-Wheeled-Humanoid-Loco-Manipulation/}{\mbox{MMR Page}}.
\end{abstract}


\section{Introduction}
Whole-body teleoperation for humanoids has established general retargeting and control procedures, but wheeled-base robots represent a topology-mismatched humanoid setting with substantial room for further study. We first need a Topology-Adaptive Motion Retargeting method that dynamically selects an effective mapping subset according to the morphological differences between the human and robot, and uses representative human-pose information for robust cross-topology retargeting. We also need a learning-based T-FeaCG for the base, which adaptively reshapes the raw upper-layer motion intent and improves transient base dynamics such as wheel--ground contact margin and actuator saturation.

In this work, we propose a whole-body teleoperation method dedicated to topology-mismatched humanoids. A wheeled base equipped with lidar SLAM provides accurate local-frame pose observations. We introduce this base-position state into the policy space, overcoming the noisy state-estimation limitation of VIO commonly used by legged robots and enabling precise base following in confined spaces. We filter the dataset using dynamic-stability constraints on the retargeted motions, removing invalid samples such as those whose center of mass leaves the support region of the base.

Our contributions are threefold: (1) a whole-body retargeting and teleoperation method adapted to topology-mismatched humanoids; (2) motion-tracking experiments in simulation that demonstrate effective imitation; and (3) an open retargeting dataset and complete codebase for R1Pro.

\section{Related Work}
Learning-based humanoid control has progressed rapidly, but most existing studies focus on legged robots \cite{h2o2024,omnih2o2024,beyondmimic2025,bfmzero2025}. Wheeled-base robots have mainly been studied with model-based control \cite{du2024efficient,l2026generativempc} or decoupled upper- and lower-body control \cite{qiao2025signbot,wang2025seec}; learning-based whole-body control remains less common because of the strongly coupled upper-body--base dynamics. Common retargeting methods \cite{luo2023perpetual,araujo2025retargeting} are mostly designed for legged robots and do not explicitly address topology-mismatched humanoids. Our method transfers retargeting to this setting with morphology-aware mapping. Four-wheel-steering bases have also been studied \cite{araujo2025motion,zhang2025hybrid}, but their mechanisms are more complex and are not readily combined with learning-based models. Pose estimation remains difficult: OmniH2O \cite{omnih2o2024} uses historical frames to improve pose estimation, and CLOT \cite{wang2026clot} corrects pose drift with high-frequency global-localization feedback. In contrast, the lidar--inertial SLAM system on our three-steer-wheel base benefits from the intrinsic constraints of planar motion and avoids the localization bias caused by gait disturbances in legged robots.

\section{Whole-Body Retargeting and Teleoperation Method}
This section describes a whole-body teleoperation system based on a Teacher--Student DAgger framework that supports inertial motion capture and VR control, together with the reward design and data-selection procedures.

\subsection{Problem Formulation}
\begin{figure*}[t]
  \centering
  \includegraphics[width=\textwidth]{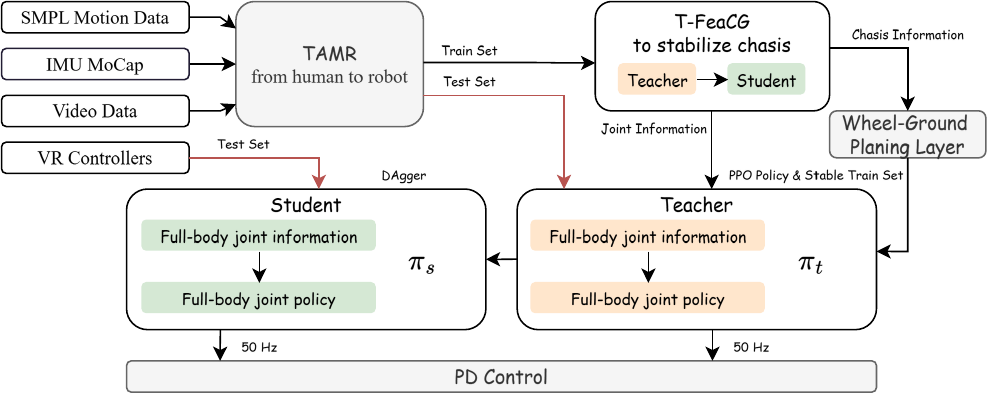}
  \caption{Whole-body teleoperation pipeline. TAMR performs retargeting, T-FeaCG adjusts base commands, and Teacher--Student DAgger performs policy training. PD control is used in simulation; the R1 Pro does not support PD control, so the PD-based results are limited to simulation.}
  \label{fig:pipeline}
\end{figure*}

We formulate the task as a Markov decision process consisting of state, action, transition, reward, and discount factor. The state contains robot proprioception and target states supplied by the human-motion input interface. The action specifies target joint angles and the base pose in $SE(2)$ expressed in the local frame. We use a simple tracking reward and train with proximal policy optimization (PPO) to maximize the discounted return. The resulting policy is organized as the three-part pipeline shown in Fig.~\ref{fig:pipeline}: TAMR performs retargeting, T-FeaCG adjusts base commands, and Teacher--Student DAgger trains the policy. The pipeline provides a general interface and a directly usable data path for wheeled-base topology-mismatched humanoids.

\subsection{Topology-Adaptive Motion Retargeting}
We use the AMASS dataset \cite{mahmood2019amass} for motion retargeting. Building on GMR \cite{araujo2025retargeting}, we design a morphology-aware Topology-Adaptive Motion Retargeting (TAMR) method for the Galaxea R1 Pro, a representative topology-mismatched humanoid. Because of its joint limits, the R1 Pro cannot simultaneously map the human leg and lower-back posture information. We therefore construct two mapping tables: a fine mapping for the human waist and back, and a coarse mapping that represents the human legs. The two-arm mapping chains use independent local scaling along three axes, together with whole-body yaw decomposition protection, to reproduce as much of the human posture as possible. We design a stable, natural, and jitter-free mapping switch activated by the human pelvis height. Since the R1 Pro cannot recover from a tip-over, we further filter the data using center-of-mass projection and zero-moment-point dynamic-stability criteria, producing a dynamically safe retargeted dataset of more than 5,000 reference motions.

\par
\begin{figure}[t]
\centering
\includegraphics[width=\columnwidth]{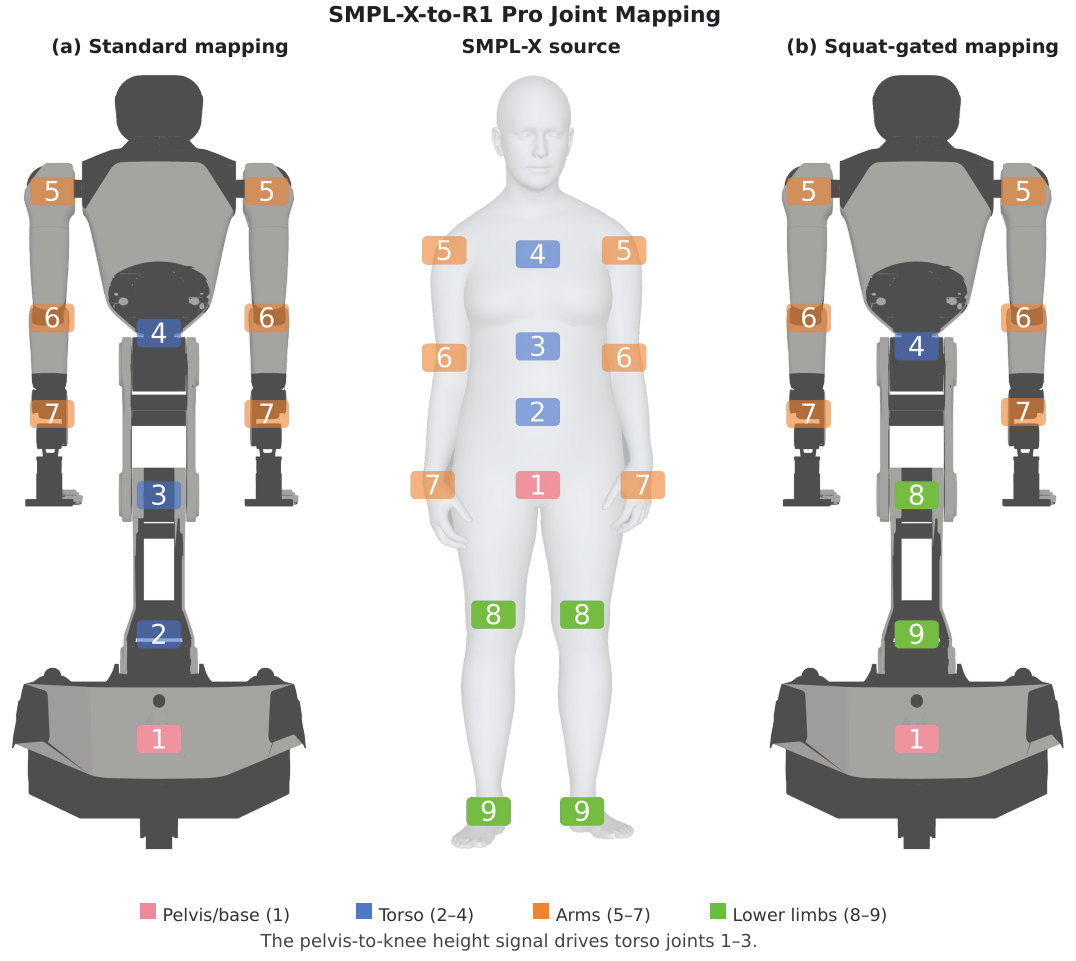}
\caption{SMPL-X-to-R1 Pro joint mapping.}
\label{fig:mapping}
\end{figure}

\noindent\textbf{TAMR Joint Mapping.} Figure~\ref{fig:mapping} shows two mappings from SMPL-X to Galaxea R1 Pro. The center is the SMPL-X source skeleton; numbers and colors denote the pelvis (1), torso (2--4), upper-limb (5--7), and lower-limb (8--9) groups. The conventional mapping on the left maps pelvis to base-link, spine1--3 to torso-link1, torso-link2, and torso-link4, and maps the corresponding shoulder, elbow, and wrist joints. The squat-gated mapping on the right retains pelvis, spine3, and upper-limb mappings. It computes the gate weight $\alpha$ from the pelvis-to-knee height difference
\begin{equation}
h=z_{\mathrm{pelvis}}-\frac{z_{\mathrm{knee,L}}+z_{\mathrm{knee,R}}}{2},
\end{equation}
and uses it to control torso-joint1--3 and imitate squatting. Knee height only triggers the gate; it is a morphological approximation and is not a strict mathematical correspondence to the robot torso joints.

\par
\begin{figure}[t]
\centering
\includegraphics[width=\columnwidth]{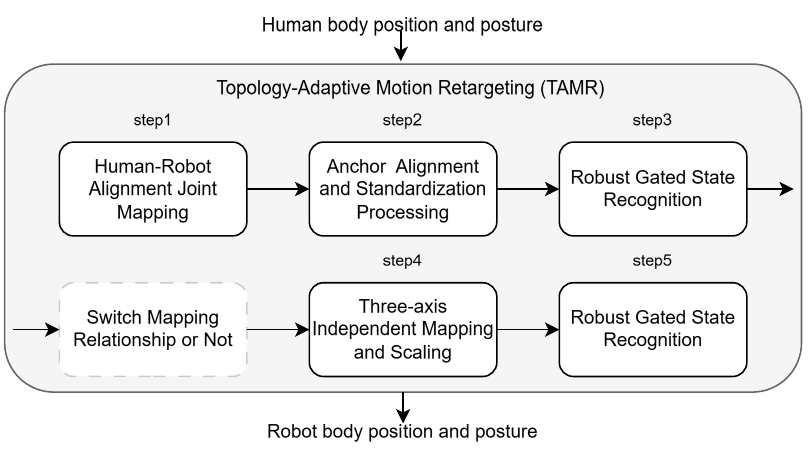}
\caption{TAMR procedure.}
\label{fig:tamr}
\end{figure}

\noindent\textbf{TAMR Procedure.} Figure~\ref{fig:tamr} summarizes the five TAMR steps. In Step 1, BVH or SMPL-X input is mapped with Standard mapping, which emphasizes fine human waist and back information. In Step 2, different SMPL-X shape parameters $\beta$ are normalized to a canonical reference shape $\beta_{\mathrm{can}}$ before forward kinematics:
\begin{equation}
\beta\leftarrow\beta_{\mathrm{can}},\qquad
p^{h}_{j,t},R^{h}_{j,t}\leftarrow\operatorname{FK}_{\mathrm{SMPLX}}(\beta_{\mathrm{can}}).
\end{equation}
Here $j$ is a joint index, $t$ is a frame index, $h$ denotes human data, and $p$ and $R$ denote joint position and rotation. The human-motion dataset is
\begin{equation}
\mathcal{H}=\left\{(p^{h}_{j,t},R^{h}_{j,t})\;\middle|\; j\in\mathcal{J}_{h},\ t\in\{1,\ldots,T\}\right\}.
\end{equation}
The robot state is
\begin{equation}
q_t^r=[x_t,y_t,\psi_t,q_t^{\mathrm{body}},q_t^{\mathrm{wheel}}],
\end{equation}
where $r$ denotes robot, $(x_t,y_t,\psi_t)$ is the planar base pose in the base frame, $q_t^{\mathrm{body}}$ contains four torso and fourteen arm joints, and $q_t^{\mathrm{wheel}}$ contains three wheel speeds and three steering angles.

For non-arm joints, mapping and scaling are
\begin{equation}
p^{\mathrm{target}}_j=p_{\mathrm{root}}+s_j(p^h_j-p^h_{\mathrm{root}}),
\end{equation}
where $s_j$ is a manually tuned scale factor and root denotes the human pelvis. For the arms, we use a parent-wise mapping:
\begin{align}
p^{\mathrm{target}}_j&=p^{\mathrm{target}}_{\pi(j)}+s_j(p^h_j-p^h_{\pi(j)}),\\
\pi(\mathrm{elbow})&=\mathrm{shoulder},\qquad
\pi(\mathrm{wrist})=\mathrm{elbow}.
\end{align}
Thus, the arms use independent local three-axis scales.

In Step 3, let $h_t$ be the current gate height and $z$ denote local-frame height. We define
\begin{equation}
h_t=z^{h}_{\mathrm{pelvis},t}-\frac{z^{h}_{\mathrm{knee,L},t}+z^{h}_{\mathrm{knee,R},t}}{2}.
\end{equation}
The raw sigmoid activation is
\begin{equation}
\alpha_t^{\mathrm{raw}}=\frac{1}{1+\exp\left(-(h_t-h_{\mathrm{mid}})/\sigma\right)},
\end{equation}
where $h_{\mathrm{mid}}$ is a preset threshold and $\sigma$ controls steepness. Low-pass filtering gives
\begin{equation}
\alpha_t=\beta\alpha_t^{\mathrm{raw}}+(1-\beta)\alpha_{t-1}.
\end{equation}
We then apply a dead-zone guard:
\begin{equation}
\rho_t=\operatorname{clip}\left(\frac{\alpha_t-\alpha_{\mathrm{guard}}}{\alpha_{\mathrm{ramp}}},0,1\right).
\end{equation}
When $\rho_t=1$ the squat-gated mapping is active; when $\rho_t=0$ Standard mapping is active. In between, torso joints are interpolated as
\begin{equation}
q_t^{\mathrm{torso}}=(1-\rho_t)q_t^{\mathrm{IK}}+\rho_tq^{\mathrm{max}}_{\mathrm{squat}}.
\end{equation}
Step 4 uses the mapped skeletal parent as the root for each arm link and assigns independent local scales along $x$, $y$, and $z$. Step 5 solves inverse kinematics in two stages: the first emphasizes end-effector orientation, and the second emphasizes whole-body position.

\subsection{Temporal Feasibility-Aware Command Governor}
Direct dynamic control of TAMR motions can cause base tip-over and unreliable wheel--ground contact. We therefore insert a regulation layer between the PPO output and the base inverse-kinematics solver. The PPO network outputs a base pose in $SE(2)$; instead of sending this pose directly to the solver, we use a Transformer-based T-FeaCG. T-FeaCG has Teacher and Student parts. The Teacher observes the intended base velocity, steering-angle jumps, minimum steering-angle error, wheel--ground contact confidence, command--response error, wheel-speed saturation, and four historical frames. It is trained using a randomly selected AMASS subset and aggressive high-transient reference trajectories to softly reshape the raw base-tracking intent. The Student omits the unavailable wheel--ground contact confidence and uses the same architecture.

After introducing T-FeaCG, steering switches, saturation, base overshoot, and oscillation are reduced, while wheel--ground contact quality improves. The governor alleviates the kinematic coupling and equivalent steering branches of multiple independently driven steer wheels. We freeze the trained Student and insert it into the PPO whole-body teleoperation pipeline to suppress unreachable aggressive commands and reduce the intent-to-actuation gap. The base inverse-kinematics layer additionally includes wheel-speed dead zones and steering-angle limits.

Our baseline uses a four-frame short history and a Transformer with 27 features per frame for supervised regression. We first compute an analytical feasible scaling from the three-wheel geometry, steering-angle errors, and maximum wheel speed, then down-weight it using contact quality and base command-response error and limit its frame-to-frame change. The two soft scaling outputs are $[\gamma_{xy},\gamma_{yaw}]$. The supervised labels combine steering reachability, wheel-speed reachability, and contact quality. After analytical post-processing, the scalar target $\gamma_b^*$ is used in
\begin{equation}
\mathcal{L}_{\mathrm{MSE}}(\theta)=\frac{1}{2B}\sum_{b=1}^{B}\sum_{j\in\{xy,\mathrm{yaw}\}}
\left(f_{\theta}(X_b)_j-\gamma_b^*\right)^2.
\label{eq:tfeacg-mse}
\end{equation}
Here, $X_b$ is a 4-frame window of 27-dimensional features and $B$ is the minibatch size. The two output dimensions and the batch samples are equally weighted. The training set contains 4,500 episodes of 100 frames generated at 50~Hz. The $x/y$ components are bounded by $\pm0.8$~m/s, the yaw rate by $\pm1.5$~rad/s, and the steering angles are initialized uniformly in $[-\pi/2,\pi/2]$ and then subjected to a bounded random walk.

\par
\begin{figure}[t]
\centering
\includegraphics[width=\columnwidth]{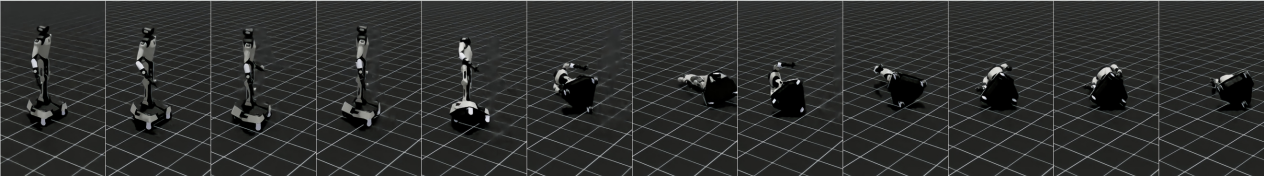}
\caption{An unsuitable TAMR base motion causes the robot to fall in simulation.}
\label{fig:fall}
\end{figure}

\begin{figure}[t]
\centering
\includegraphics[width=\columnwidth]{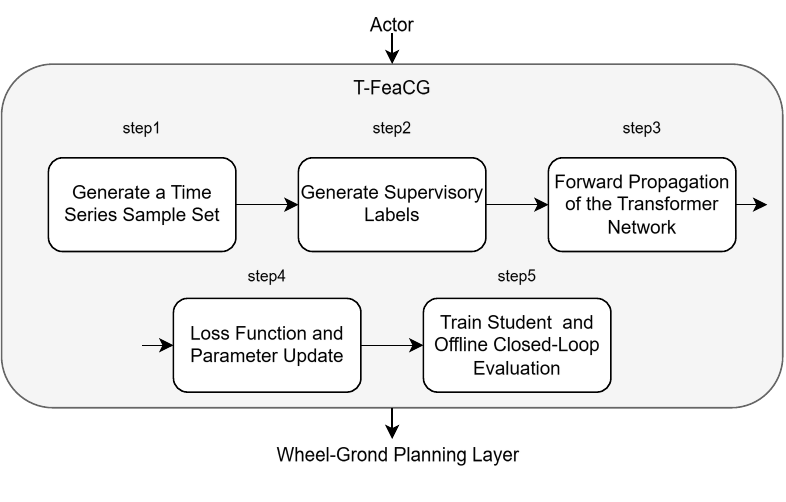}
\caption{T-FeaCG architecture. Steps 1--4 belong to the Teacher and Step 5 to the Student.}
\label{fig:govarch}
\end{figure}

\noindent\textbf{T-FeaCG Architecture.} Direct dynamic control of TAMR can tip over the base or produce insufficient wheel--ground contact, as shown in Fig.~\ref{fig:fall}. T-FeaCG is a supervised-distillation framework with Teacher and Student parts (Fig.~\ref{fig:govarch}) that uses a label $\gamma^*$ to scale translational and yaw intent. The fall in Fig.~\ref{fig:fall} is caused by an unstable TAMR base trajectory combined with a strong base-tracking reward.

The Teacher's contact-quality input comes from simulated contact forces and is unavailable on the physical robot. Because this quantity is important for tip-over prevention and intent safety, we define a score in simulation and roll out a large dataset to fit it. The Student omits only three input dimensions and is trained with behavior cloning.

For each episode, Step 1 samples an autoregressive base-twist sequence, random initial steering angles, and random contact-quality dropouts. Inverse kinematics maps the raw intent $v^{\mathrm{intent}}=[v_x,v_y,\omega_z]$ to steering angles $\theta_i$ and wheel-center speeds $u_i$. Simulated delay and Gaussian noise produce $v^{\mathrm{measured}}$, with residual
\begin{align}
e_v&=v^{\mathrm{intent}}-v^{\mathrm{measured}},\\
e_{v,xy}&=[e_{v,x},e_{v,y}],\\
e_{\omega_z}&=\omega_z^{\mathrm{intent}}-\omega_z^{\mathrm{measured}}.
\end{align}
Hysteresis and first-order low-pass filtering yield $q_i\in[0,1]$. The raw feature vector $s_t\in\mathbb{R}^{27}$ contains steering error, wheel-speed demand, contact confidence, and velocity residual; the temporal window is
\begin{equation}
S_t=[s_{t-3},s_{t-2},s_{t-1},s_t]\in\mathbb{R}^{4\times27}.
\end{equation}
Step 2 computes steering reachability
\begin{equation}
g_s=\operatorname{clip}\left(\frac{\dot\theta_{\max}T_p}{\max_i e_{\theta,i}},0.25,1\right),
\end{equation}
and wheel-speed reachability
\begin{equation}
g_w=\min\left(1,\frac{\dot q_{\max}r}{\max_i\lVert u_i\rVert}\right).
\end{equation}
We define
\begin{equation}
g_{\mathrm{raw}}=\min(g_s,g_w),\quad c=0.35+0.65\min_iq_i,
\end{equation}
\begin{align}
r_T&=\operatorname{clip}\left(1-0.22\lVert e_{v,xy}\rVert^2-0.08|e_{\omega_z}|,0.4,1\right),\\
g_{\mathrm{proposed}}&=g_{\mathrm{raw}}cr_T.
\end{align}
The smoothed label is
\begin{equation}
\gamma_t^*=\operatorname{clip}(g_{\mathrm{proposed}},\gamma_{t-1}^*-0.10,\gamma_{t-1}^*+0.10),\quad \gamma_t^*\in[0,1].
\end{equation}
Step 3 feeds $S_t$ to the network and predicts $\gamma_t$. Step 4 uses the MSE loss in (\ref{eq:tfeacg-mse}) to obtain $\theta^*$. In Step 5, behavior cloning trains the Student, which is inserted before the Wheel--Ground Planning Layer. It computes $\gamma_t=f_{\theta^*}(S_t)$ and scales the twist as
\begin{align}
v_{x}^{\mathrm{cmd}}&=\gamma_{xy}v_x^{\mathrm{intent}},\\
v_{y}^{\mathrm{cmd}}&=\gamma_{xy}v_y^{\mathrm{intent}},\\
\omega_z^{\mathrm{cmd}}&=\gamma_{yaw}\omega_z^{\mathrm{intent}}.
\end{align}


\par
\begin{figure}[t]
\centering
\includegraphics[width=\columnwidth]{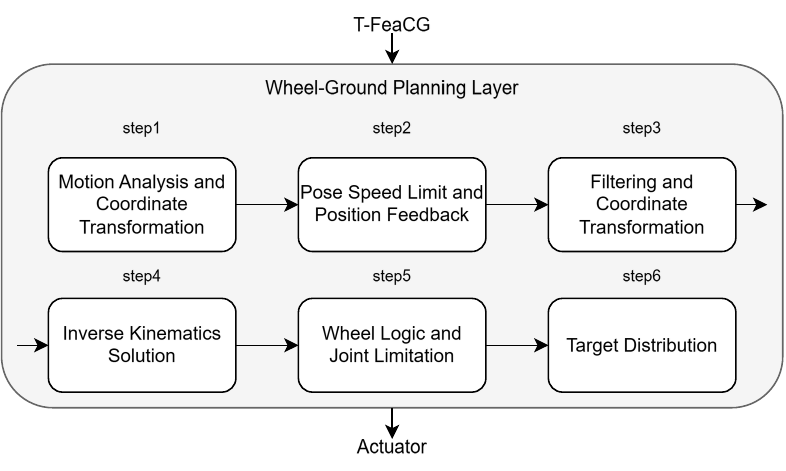}
\caption{Wheel--Ground Planning Layer.}
\label{fig:planning}
\end{figure}

\noindent\textbf{Wheel--Ground Planning Layer.} Figure~\ref{fig:planning} shows the Wheel--Ground Planning Layer. Step 1 parses the three-dimensional normalized base action from T-FeaCG, applies scale factors, and maps it to a virtual base pose in the world frame relative to the beginning of the motion segment. Step 2 applies a single-step displacement constraint and combines position feedback with reference-trajectory velocity feedforward. Let $(p_t^w,\psi_t^w)$ be the virtual target pose, $(p_{t-1}^{\mathrm{cmd}},\psi_{t-1}^{\mathrm{cmd}})$ the cached previous target, and $(\Delta p_{\mathrm{step}},\Delta\psi_{\mathrm{step}})$ the displacement limits:
\begin{align}
e_t^{\mathrm{req}}&=p_t^w-p_{t-1}^{\mathrm{cmd}}, & e_{\psi,t}^{\mathrm{req}}&=\operatorname{wrap}(\psi_t^w-\psi_{t-1}^{\mathrm{cmd}}),\\
e_t^{\mathrm{clip}}&=\operatorname{clip}(e_t^{\mathrm{req}},\pm\Delta p_{\mathrm{step}}), & e_{\psi,t}^{\mathrm{clip}}&=\operatorname{clip}(e_{\psi,t}^{\mathrm{req}},\pm\Delta\psi_{\mathrm{step}}),\\
p_t^{\mathrm{cmd}}&=p_{t-1}^{\mathrm{cmd}}+e_t^{\mathrm{clip}}, & \psi_t^{\mathrm{cmd}}&=\psi_{t-1}^{\mathrm{cmd}}+e_{\psi,t}^{\mathrm{clip}}.
\end{align}
For the measured pose $(p_t^{\mathrm{robot}},\psi_t^{\mathrm{robot}})$, feedback gains $K_p,k_\psi$, and correction limits $\delta v_{\mathrm{lim}},\delta\omega_{\mathrm{lim}}$, we compute
\begin{align}
\Delta p_t&=p_t^{\mathrm{cmd}}-p_t^{\mathrm{robot}},\\
\Delta\psi_t&=\operatorname{wrap}(\psi_t^{\mathrm{cmd}}-\psi_t^{\mathrm{robot}}),\\
\delta v_t&=\operatorname{clip}(K_p\odot\Delta p_t,\pm\delta v_{\mathrm{lim}}),\\
\delta\omega_t&=\operatorname{clip}(k_\psi\Delta\psi_t,\pm\delta\omega_{\mathrm{lim}}),\\
v_{t}^{w,\mathrm{raw}}&=v_{t}^{\mathrm{ref},w}+\delta v_t,\\
\omega_{z,t}^{w,\mathrm{raw}}&=\omega_{z,t}^{\mathrm{ref},w}+\delta\omega_t.
\end{align}
Step 3 applies EMA filtering and transforms the command to the body frame:
\begin{align}
v_t^{w,f}&=\alpha_{\mathrm{EMA}}v_t^{w,\mathrm{raw}}+(1-\alpha_{\mathrm{EMA}})v_{t-1}^{w,f},\\
\omega_{z,t}^{w,f}&=\alpha_{\mathrm{EMA}}\omega_{z,t}^{w,\mathrm{raw}}+(1-\alpha_{\mathrm{EMA}})\omega_{z,t-1}^{w,f},\\
v_t^b&=R(\psi_t^{\mathrm{robot}})^T v_t^{w,f}, & \omega_{z,t}^b&=\omega_{z,t}^{w,f}.
\end{align}
Acceleration limits give
\begin{align}
v_{t}^{b,\mathrm{lim}}&=v_{t-1}^{b}+\operatorname{clip}(v_t^b-v_{t-1}^b,\pm\Delta v_{b,\max}),\\
\omega_{z,t}^{b,\mathrm{lim}}&=\omega_{z,t-1}^{b}+\operatorname{clip}(\omega_{z,t}^b-\omega_{z,t-1}^b,\pm\Delta\omega_{b,\max}).
\end{align}
Step 4 solves the omnidirectional steer-wheel inverse kinematics. For wheel position $(x_i,y_i)$,
\begin{align}
v_{x,i}&=v_{x,t}^{b,\mathrm{lim}}-\omega_{z,t}^{b,\mathrm{lim}}y_i, & v_{y,i}&=v_{y,t}^{b,\mathrm{lim}}+\omega_{z,t}^{b,\mathrm{lim}}x_i,\\
\theta_i^{\mathrm{raw}}&=\operatorname{atan2}(v_{y,i},v_{x,i}), & \phi_i^{\mathrm{raw}}&=\frac{\sqrt{v_{x,i}^2+v_{y,i}^2}}{r_{\mathrm{wheel}}}.
\end{align}
Step 5 handles low-speed hysteresis and steering-angle ambiguity. For $u_i=\sqrt{v_{x,i}^2+v_{y,i}^2}$,
\begin{align}
u_i\ge v_{\mathrm{enter}} &: (\theta_i,\phi_i)\leftarrow(\theta_i^{\mathrm{raw}},\phi_i^{\mathrm{raw}}),\\
u_i\le v_{\mathrm{exit}} &: \theta_i\leftarrow\theta_{i,t-1},\quad \phi_i\leftarrow0.
\end{align}
The two physically equivalent solutions $(\theta_i,\phi_i)$ and $(\theta_i+\pi,-\phi_i)$ are resolved by selecting the one closest to the previous steering angle. Joint-rate and amplitude limits are then applied, and Step 6 sends the three steering-position and three wheel-speed targets to the actuators.

\subsection{Teacher--Student DAgger}
Our reinforcement-learning framework follows the network designs of OmniH2O \cite{omnih2o2024} and BeyondMimic \cite{beyondmimic2025}. We introduce a Teacher--Student DAgger architecture in which inertial motion capture and VR can connect directly to the Teacher and Student, respectively. The Teacher uses an asymmetric Actor--Critic architecture. The Actor observes joint angles and velocities, the $SE(2)$ base pose and velocity from lidar SLAM, and the retargeted target pose. The Critic additionally receives privileged information such as the absolute local-frame pose of every robot joint. After T-FeaCG training, the governor is frozen and the whole-body policy is trained with PPO. The Student targets VR operation: sparse keypoint poses from a head-mounted display and two hand controllers are retargeted to a robot base $SE(2)$ pose intent and two wrist target configurations.

Compared with OmniH2O, our observation includes local-frame $SE(2)$ and velocity but no historical policy input. We use lidar SLAM because it provides sufficiently accurate localization and simplifies training. We train a deployable policy with DAgger \cite{ross2010dagger}. In addition to tracking rewards, we add a wheel--ground contact-time reward. Simulation shows that this concise reward is sufficient for good tracking. We apply small domain randomization to friction, center-of-mass offset, proportional gain, and derivative gain. The Student is an MLP with hidden widths $512\rightarrow256\rightarrow128$ and output width 21. In each DAgger round, 4,500 motions generate Student observations $o_t^S$ and Teacher labels $a_t^T$ in simulation. Across ten rounds, $\beta$ decreases linearly from 0.25 to zero, and the mixed action in round $k$ is
\begin{equation}
a_t^{\mathrm{exec}}=\operatorname{clip}\left(\beta_k a_t^T+(1-\beta_k)a_t^S,-1,1\right).
\label{eq:dagger-mix}
\end{equation}
The replay buffer contains 20,000,000 rows, with 30\% reserved for states with large Student--Teacher action differences. Each DAgger iteration first samples one motion-set round in simulation and then trains once on the accumulated replay buffer using the 21-dimensional Teacher--Student action MSE.

The PPO hyperparameters, Actor observations, Actor actions, and rewards are listed in Tables~\ref{tab:ppo},~\ref{tab:obs},~\ref{tab:act}, and~\ref{tab:reward}. The Actor and Critic use hidden layers $[512,256,128]$ and ELU activation. The action distribution is squashed Gaussian; the upper action initialization standard deviation is 0.25, the base action initialization standard deviation is 0.02, the entropy coefficient is 0.005, the learning rate is 0.0003, rollout length is 24 steps per environment per iteration, PPO uses 3 epochs and 4 minibatches, $\gamma=0.99$, $\lambda=0.95$, maximum gradient norm is 1, and control frequency is 50~Hz.

\begin{table}[H]
\centering
\caption{PPO policy and training hyperparameters.}
\label{tab:ppo}
\begin{tabular}{ll}
\toprule
Parameter & Value\\\midrule
Actor/Critic hidden layers & $[512,256,128]$\\
Activation & ELU\\
Action distribution & Squashed Gaussian\\
Upper action init. std. & 0.25\\
Base action init. std. & 0.02\\
Entropy coefficient & 0.005\\
Learning rate & 0.0003\\
Rollout steps/env/iter. & 24\\
PPO epochs/minibatches & 3/4\\
$\gamma,\lambda$ & 0.99, 0.95\\
Maximum gradient norm & 1\\
Control period & 50~Hz\\
\bottomrule
\end{tabular}
\end{table}

\begin{table}[H]
\centering
\caption{Actor observations. All observations are sensor-accessible in the asymmetric Actor--Critic architecture.}
\label{tab:obs}
\resizebox{\columnwidth}{!}{\begin{tabular}{lcl}
\toprule
Term & Dim. & Meaning\\\midrule
command & 48 & 24 joint angles and velocities of the reference motion\\
motion\_anchor\_b & 9 & Anchor position error and relative orientation\\
base\_vel & 6 & Measured root linear and angular velocity\\
joint & 48 & Measured 24 joint angles and velocities\\
last\_actions & 21 & Previous action\\
motion\_pose & 3 & Reference-motion pose\\
robot\_pose & 3 & Measured robot pose\\\midrule
Total & 138 & \\
\bottomrule
\end{tabular}}
\end{table}

\begin{table}[H]
\centering
\caption{Actor action output.}
\label{tab:act}
\begin{tabular}{lcl}
\toprule
Term & Dim. & Meaning\\\midrule
Base target & 3 & Virtual base target\\
Torso & 4 & Normalized torso-joint position command\\
Left arm & 7 & Normalized left-arm position command\\
Right arm & 7 & Normalized right-arm position command\\
\bottomrule
\end{tabular}
\end{table}

The action vector is
\begin{equation}
a_t=[a_x,a_y,a_\psi,a_{\mathrm{torso},1:4},a_{\mathrm{left},1:7},a_{\mathrm{right},1:7}]\in[-1,1]^{21}.
\end{equation}
For upper-body joint $i$, define
\begin{equation}
s_i^-=0.95(q_{0,i}-q_{i}^{\min}),\qquad s_i^+=0.95(q_i^{\max}-q_{0,i}),
\end{equation}
and map the action to the requested joint angle by
\begin{equation}
q_{i,t}^{\mathrm{req}}=\begin{cases}
q_{0,i}+a_{i,t}s_i^+, & a_{i,t}\ge0,\\
q_{0,i}+a_{i,t}s_i^-, & a_{i,t}<0.
\end{cases}
\end{equation}
Here $q_{0,i}$ is the default joint angle and $q_i^{\min},q_i^{\max}$ are the joint limits.

\begin{table*}[!t]
\centering
\caption{Reward terms. The total reward is $R_t=\sum_k w_k r_{k,t}$.}
\label{tab:reward}
\small
\setlength{\tabcolsep}{4pt}
\begin{tabularx}{\textwidth}{>{\raggedright\arraybackslash}p{4.2cm}X p{1.0cm}}
\toprule
Reward term & Formula & Weight\\
\midrule
Anchor position & $r_{\mathrm{anchor,pos}}=\exp\!\left(-\frac{\lVert p_a^{\mathrm{ref}}-p_a\rVert_2^2}{0.3^2}\right)$ & 0.5\\
Anchor orientation & $r_{\mathrm{anchor,ori}}=\exp\!\left(-\frac{\theta(q_a^{\mathrm{ref}},q_a)^2}{0.4^2}\right)$ & 0.5\\
Upper-body relative position & $r_{\mathrm{body,pos}}=\exp\!\left(-\frac{1}{0.6^2}\frac{1}{|\mathcal U_b|}\sum_{b\in\mathcal U_b}\lVert p_{b,\mathrm{rel}}^{\mathrm{ref}}-p_{b,\mathrm{rel}}\rVert_2^2\right)$ & 1.0\\
Upper-body relative orientation & $r_{\mathrm{body,ori}}=\exp\!\left(-\frac{1}{0.8^2}\frac{1}{|\mathcal U_b|}\sum_{b\in\mathcal U_b}\theta(q_{b,\mathrm{rel}}^{\mathrm{ref}},q_{b,\mathrm{rel}})^2\right)$ & 1.0\\
Body linear-velocity tracking & $r_{\mathrm{body,linvel}}=\exp\!\left(-\frac{1}{1.0^2}\frac{1}{|\mathcal B|}\sum_{b\in\mathcal B}\lVert v_{b,w}^{\mathrm{ref}}-v_{b,w}\rVert_2^2\right)$ & 1.0\\
Body angular-velocity tracking & $r_{\mathrm{body,angvel}}=\exp\!\left(-\frac{1}{\pi^2}\frac{1}{|\mathcal B|}\sum_{b\in\mathcal B}\lVert\omega_{b,w}^{\mathrm{ref}}-\omega_{b,w}\rVert_2^2\right)$ & 1.0\\
Upper-body joint angle & $r_{\mathrm{joint,focus}}=-\frac{1}{|\mathcal U_j|}\sum_{j\in\mathcal U_j}e_{q,j}^2$ & 1.0\\
Base pose tracking & $r_{\mathrm{base,pose}}=-(e_{xy}^2+e_\psi^2),\quad e_{xy}=\lVert p_{\mathrm{base},xy}^{\mathrm{ref}}-p_{\mathrm{base},xy}\rVert_2,\quad e_\psi=\operatorname{wrap}(\psi^{\mathrm{ref}}-\psi)$ & 1.0\\
Regularization & $r_{\mathrm{rate}}=\sum_{k=1}^{21}(a_{k,t}-a_{k,t-1})^2$ & $-0.3$\\
\bottomrule
\end{tabularx}
\end{table*}
\FloatBarrier

\section{Experimental Results}
We analyze TAMR, T-FeaCG, and Teacher--Student DAgger. Some test videos are released on the project website.

\subsection{TAMR Results}
TAMR is designed for topology-mismatched humanoids, whereas the GMR baseline is better suited to humanoids such as Unitree G1. A fair comparison on one robot morphology is therefore unavailable. We report the output poses and metrics side by side on their respective robots and perform ablations. We select 20 AMASS motions and repeat each ten times. Joint errors are averaged within each motion and then across all motions. We measure torso position and orientation errors with respect to the ground, wrist position error with respect to the shoulder, and wrist orientation error with respect to the ground. GMR uses G1; TAMR uses the R1 Pro. The variants TAMR$_{\mathrm{w/o\ gate}}$, TAMR$_{\mathrm{w/o\ norm}}$, and TAMR$_{\mathrm{root-base}}$ remove robust switching, morphology normalization, and shoulder-based arm scaling, respectively. Errors are reported in centimeters and degrees. The switching transient velocity $v_{\mathrm{switch}}$ is the instantaneous Cartesian velocity magnitude of feature points around a TAMR mapping switch; lower values indicate a smoother switch.

\begin{figure}[t]
  \centering
  \includegraphics[width=\columnwidth]{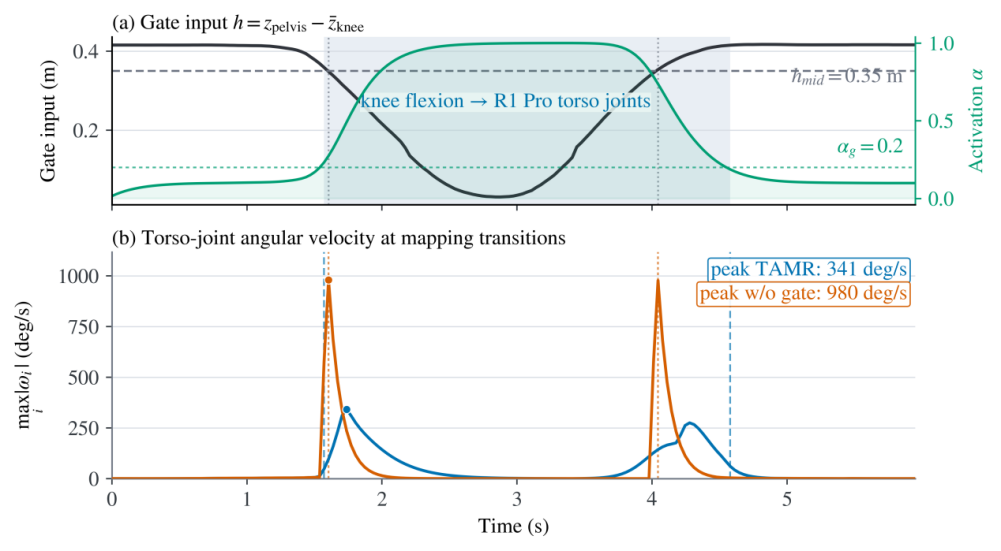}
  \caption{Mapping-switch and torso-angular-velocity responses. The black curve is the gate input, the green curve is the smoothed activation, the blue curve is the robust-gate response, and the orange curve is the hard-switch ablation.}
  \label{fig:switch}
\end{figure}

Table~\ref{tab:tamr} summarizes the quantitative results. Robust gating suppresses switching transients, shoulder-based arm scaling reduces wrist--shoulder position error, and morphology normalization provides a relatively small gain. Figure~\ref{fig:switch} shows a squat-to-stand example; the robust gate substantially reduces the torso joint angular-velocity peak.

\begin{table}[t]
\centering
\caption{Simulation of TAMR on the selected dataset. GMR is evaluated on G1 and the other variants on R1 Pro. GMR is a reference only and is not ranked against TAMR. Its switching velocity is unavailable.}
\label{tab:tamr}
\resizebox{\columnwidth}{!}{\begin{tabular}{lccccc}
\toprule
Method & $e_{p,\mathrm{torso}}$ & $e_{R,\mathrm{torso}}$ & $e_{p,\mathrm{wrist-shoulder}}$ & $e_{R,\mathrm{wrist}}$ & $v_{\mathrm{switch}}$\\
\midrule
GMR & 3.464 & 7.342 & 14.775 & 2.748 & --\\
TAMR & 3.684 & 12.968 & 8.788 & 2.689 & 0.053\\
TAMR$_{\mathrm{w/o\ gate}}$ & 3.682 & 12.862 & 8.772 & 3.055 & 0.817\\
TAMR$_{\mathrm{w/o\ norm}}$ & 3.748 & 12.985 & 10.090 & 2.720 & 0.413\\
TAMR$_{\mathrm{root-base}}$ & 3.831 & 14.334 & 93.562 & 5.636 & 0.502\\
\bottomrule
\end{tabular}}
\end{table}
\FloatBarrier

\FloatBarrier
\subsection{T-FeaCG Results}
\begin{figure}[t]
  \centering
  \includegraphics[width=\columnwidth]{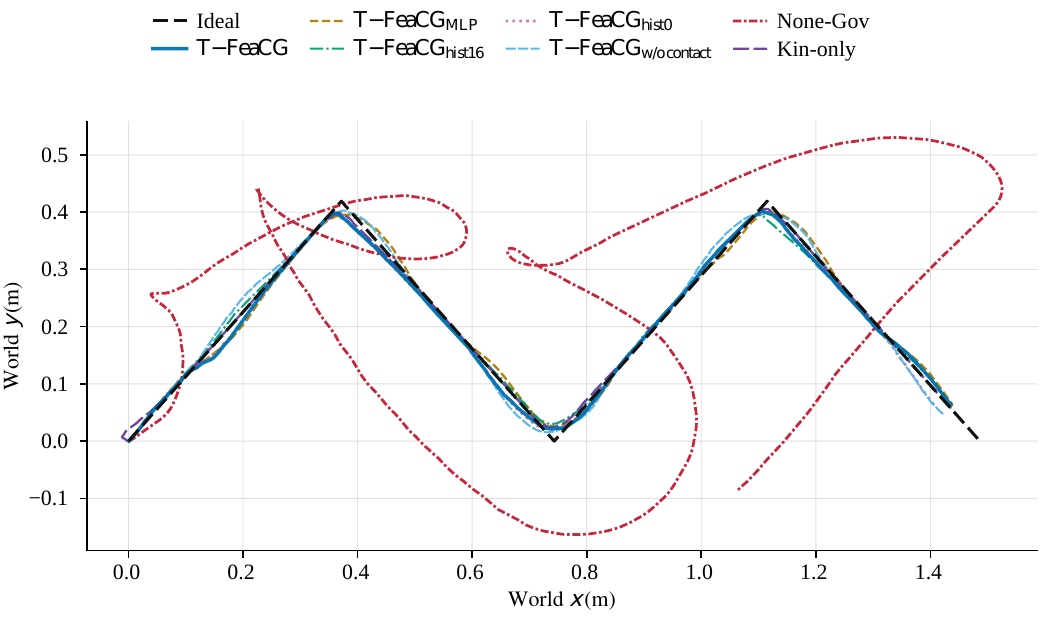}
  \caption{T-FeaCG ablation curves on an M-shaped trajectory. ``Ideal'' is the target trajectory. T-FeaCG has a base-position error of 7.41~cm in this replay, the second-lowest value; Kin-only is lower at 5.31~cm.}
  \label{fig:tfeacg}
\end{figure}

We evaluate the governor with several ablations: replacing the Transformer with an MLP; replacing the four-frame history with 16 or zero frames; removing contact confidence; removing T-FeaCG entirely; and using a kinematics-only base. The test set contains aggressive O-, M-, S-, V-, and L-shaped trajectories with sharp turns and rapid speed changes. Each experiment is repeated ten times. We evaluate local-frame base $SE(2)$ error, the fraction of time all three wheels remain in contact, the mean number of steering switches, and the fraction of time at the steering limit. Figure~\ref{fig:tfeacg} shows one M-shaped replay.

Table~\ref{tab:tfeacg} shows that T-FeaCG reduces steering switches and time at the steering limit. Four history frames provide the best balance between performance and learning efficiency. The Transformer performs better than the MLP in the overall evaluation. The test paths average 8~m, the mean speed is 0.1268~m/s, the control period is 0.02~s, and the steering limit is $\pm180^\circ$. The intrinsic turning-radius constraint of the three-steer-wheel R1 Pro makes sharp-corner tracking challenging; T-FeaCG provides useful regulation in this setting.

\begin{table}[!htbp]
\centering
\caption{Simulation of T-FeaCG on the selected test set.}
\label{tab:tfeacg}
\resizebox{\columnwidth}{!}{\begin{tabular}{lccccc}
\toprule
Method & $e_{p,\mathrm{base}}$ & $e_{R,\mathrm{base}}$ & $\tau_{\mathrm{contact}}$ & $N_{\mathrm{switch}}$ & $\tau_{\mathrm{sat}}$\\
\midrule
T-FeaCG & $0.0516\pm0.0023$ & $32.21\pm0.63$ & $0.361\pm0.037$ & $21.9\pm2.5$ & $0.665\pm0.032$\\
T-FeaCG$_{\mathrm{MLP}}$ & $0.0495\pm0.0016$ & $31.94\pm1.77$ & $0.326\pm0.044$ & $25.4\pm2.2$ & $0.691\pm0.011$\\
T-FeaCG$_{\mathrm{hist16}}$ & $0.0545\pm0.0012$ & $32.74\pm0.41$ & $0.350\pm0.039$ & $22.2\pm2.0$ & $0.696\pm0.024$\\
T-FeaCG$_{\mathrm{hist0}}$ & $0.0614\pm0.0018$ & $33.04\pm0.59$ & $0.0364\pm0.042$ & $22.9\pm3.9$ & $0.663\pm0.030$\\
T-FeaCG$_{\mathrm{w/o\ contact}}$ & $0.0511\pm0.0004$ & $32.79\pm0.34$ & $0.346\pm0.037$ & $24.9\pm2.2$ & $0.684\pm0.012$\\
None-Gov & $0.2784\pm0.0568$ & $39.87\pm5.54$ & $0.094\pm0.028$ & $40.4\pm11.5$ & $0.941\pm0.040$\\
Kin-only & $0.0474\pm0.0026$ & $23.45\pm0.24$ & $0.316\pm0.039$ & $79.4\pm11.8$ & $0.743\pm0.148$\\
\bottomrule
\end{tabular}}
\end{table}
\FloatBarrier

\subsection{Teacher--Student DAgger Results}

OmniH2O provides full privileged states during Teacher training but removes explicit root-frame position and velocity from the Student, relying on temporal history to encode motion and avoid VIO drift in legged robots. Our wheeled robot uses lidar SLAM and obtains a stable, low-drift local $SE(2)$ pose, so explicit root observations can be introduced safely.

We conduct Teacher and Student comparisons on the TAMR motion dataset and use the same motions and evaluation procedure as in the TAMR experiment. For the Teacher comparison, we compare our policy with T-FeaCG against the original OmniH2O Teacher to evaluate the upper-layer tracking gain. For the Student comparison, ``Ours'' uses T-FeaCG and an explicit $SE(2)$ root observation. We also evaluate Ours+Hist, Ours without T-FeaCG, OmniH2O+T, OmniH2O, H2O+T, and H2O. ``+T'' means that T-FeaCG is used in both Teacher and Student stages.

\begin{figure}[!htbp]
  \centering
  \includegraphics[width=\columnwidth]{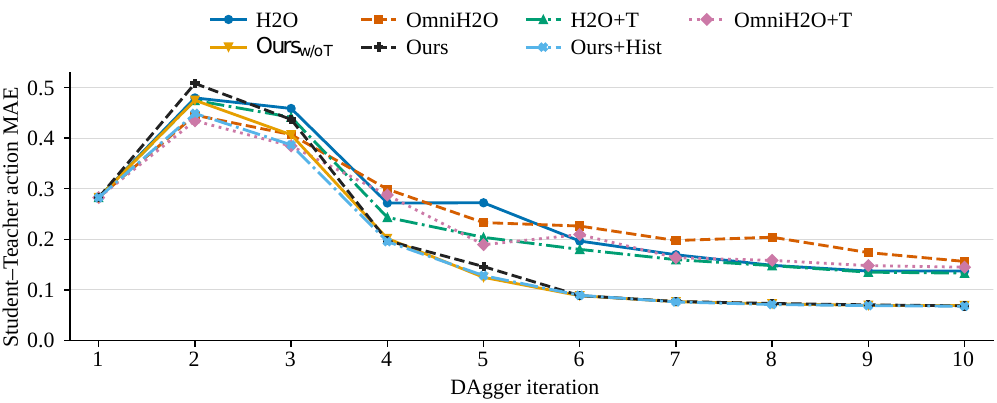}
  \caption{DAgger iteration and error. The normalized action MAE of Ours decreases from 0.283 to 0.068 in round 10, a 75.8\% reduction.}
  \label{fig:dagger}
\end{figure}

Table~\ref{tab:teacher} shows that T-FeaCG reduces torso position and orientation errors and increases contact time, while its effect on wrist errors is small. OmniH2O was designed for legged robots and its observation, action, and reward spaces are not adapted to a wheeled-base $SE(2)$. Explicit lidar-SLAM pose enables more accurate wheeled-humanoid tracking. Figure~\ref{fig:dagger} reports the DAgger-round error curve. Table~\ref{tab:student} shows that Ours reduces both base and wrist tracking errors relative to H2O and OmniH2O. Removing T-FeaCG increases the error, while adding extra history to Ours does not yield a further gain.

\begin{table}[!htbp]
\centering
\caption{Teacher-stage simulation results on the selected dataset.}
\label{tab:teacher}
\resizebox{\columnwidth}{!}{\begin{tabular}{lccccc}
\toprule
Method & $e_{p,\mathrm{torso}}$ & $e_{R,\mathrm{torso}}$ & $e_{p,\mathrm{wrist-shoulder}}$ & $e_{R,\mathrm{wrist}}$ & $\tau_{\mathrm{contact}}$\\
\midrule
Ours & $17.7742\pm6.5315$ & $19.9015\pm4.0865$ & $9.3795\pm2.9292$ & $17.5196\pm6.9986$ & $0.3137\pm0.1220$\\
OmniH2O & $20.2811\pm10.7550$ & $24.5221\pm11.9598$ & $9.2609\pm2.7724$ & $18.6244\pm6.9064$ & $0.2943\pm0.1381$\\
\bottomrule
\end{tabular}}
\end{table}

\FloatBarrier

\subsection{Joint-Jitter Analysis for Mobile-Base Coupling}
\begin{figure*}[!t]
\centering
\begin{minipage}[t]{0.48\textwidth}
\centering
\includegraphics[width=\linewidth]{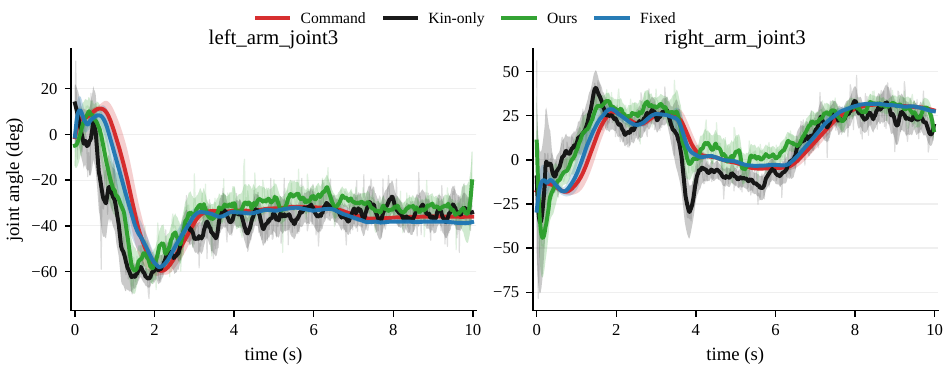}
\captionof{figure}{Actor target commands for the third joints of the left and right arms.}
\label{fig:actor-jitter}
\end{minipage}\hfill
\begin{minipage}[t]{0.48\textwidth}
\centering
\includegraphics[width=\linewidth]{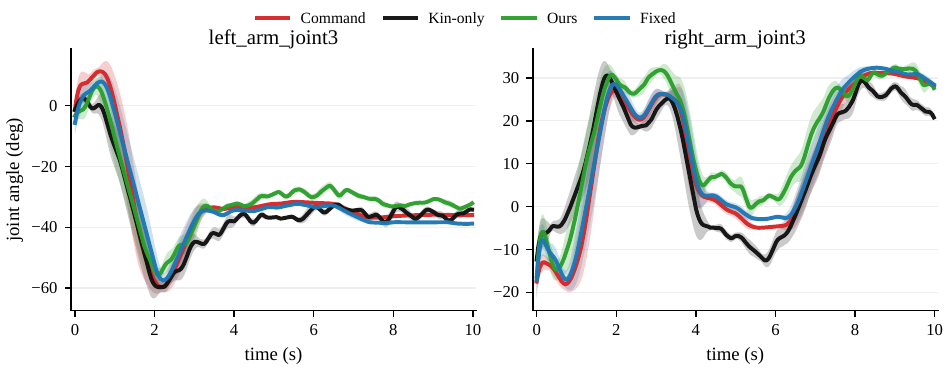}
\captionof{figure}{Actual responses of the PD controller.}
\label{fig:pd-response}
\end{minipage}
\end{figure*}

During simulation, we observed substantial noise in the Actor outputs during reinforcement-learning training and conducted a comparative experiment to identify its source. We selected the same motion sequence and analyzed the third joints of the left and right arms. Figure~\ref{fig:actor-jitter} shows the target commands sent from the Actor to the controller, whereas Fig.~\ref{fig:pd-response} shows the corresponding controller responses. ``Command'' denotes the retargeted reference command. ``Kin-only'' solves the base inverse kinematics directly without T-FeaCG or the Wheel--Ground Planning Layer. ``Fixed'' modifies the network input to train only the upper-body actions, with the base fixed in place. ``Ours'' is our baseline.

\begin{table*}[!t]
\centering
\small
\caption{Student-stage simulation results on the selected dataset.}
\label{tab:student}
\resizebox{\textwidth}{!}{\begin{tabular}{lccccc}
\toprule
Method & $e_{p,\mathrm{torso}}$ & $e_{R,\mathrm{torso}}$ & $e_{p,\mathrm{wrist-shoulder}}$ & $e_{R,\mathrm{wrist}}$ & $\tau_{\mathrm{contact}}$\\
\midrule
Ours & $15.4217\pm3.6106$ & $20.0401\pm4.3414$ & $9.8813\pm2.4137$ & $19.5877\pm2.6872$ & $0.4083\pm0.1283$\\
Ours+Hist & $16.6651\pm4.1717$ & $22.4685\pm7.1450$ & $10.3724\pm2.9062$ & $19.5431\pm2.2520$ & $0.3532\pm0.0760$\\
Ours$_{\mathrm{w/o\ T}}$ & $19.8095\pm15.1643$ & $22.6743\pm7.7755$ & $9.2709\pm1.4005$ & $20.2119\pm6.5541$ & $0.3273\pm0.1164$\\
OmniH2O+T & $61.0191\pm15.3287$ & $25.1610\pm12.9311$ & $18.9711\pm14.0111$ & $44.6633\pm39.6686$ & $0.5587\pm0.1221$\\
OmniH2O & $55.5325\pm13.0953$ & $26.8529\pm7.1242$ & $14.0634\pm4.2076$ & $38.8905\pm23.3874$ & $0.5736\pm0.1305$\\
H2O+T & $54.9903\pm9.4414$ & $25.3207\pm12.2642$ & $14.6278\pm6.0418$ & $34.1106\pm21.3270$ & $0.4264\pm0.1511$\\
H2O & $58.1512\pm13.1681$ & $29.9782\pm11.5572$ & $14.2982\pm6.2074$ & $37.8514\pm24.3684$ & $0.4755\pm0.1522$\\
\bottomrule
\end{tabular}}
\end{table*}

Figure~\ref{fig:actor-jitter} shows that Fixed is the smoothest, Ours exhibits more jitter, and Kin-only exhibits the most jitter. This indicates that the jitter originates from base control. T-FeaCG alleviates abrupt base-command changes and thereby smooths the Actor output. The base motion can make the Actor output nonsmooth because the wheeled base has unreachable motions and cannot fully track some commands, which can destabilize the base; this is also the problem addressed by T-FeaCG. These results further show that preserving motion quality when retargeting human motion to a wheeled-base robot is difficult and may require sacrificing Actor-output precision. Figure~\ref{fig:pd-response} shows the actual response after configuring the controller with a damping ratio of 2, i.e., an overdamped response. Compliant control filters the high-frequency jitter in Fig.~\ref{fig:actor-jitter}. Table~\ref{tab:jitter-rms} quantifies the results: Ours is better than Kin-only, and the PD controller reduces the RMS.

\begin{table}[!t]
\centering
\caption{High-frequency residual RMS after detrending each joint sequence with a 21-frame Savitzky--Golay filter.}
\label{tab:jitter-rms}
\resizebox{\columnwidth}{!}{\begin{tabular}{lcccc}
\toprule
Method & Left arm joint 3 & & Right arm joint 3 & \\
 & Actor & Response & Actor & Response\\
\midrule
Ours & 5.128 & 0.319 & 5.424 & 0.363\\
Kin-only & 6.567 & 0.347 & 8.093 & 0.308\\
Fixed & 0.639 & 0.089 & 0.922 & 0.119\\
\bottomrule
\end{tabular}}
\end{table}

\FloatBarrier
\vspace{-8pt}
\section{Limitations and Future Work}
\textbf{Summary:} We propose TAMR for retargeting motions to wheeled-base humanoids, T-FeaCG for improving base stability, and a Teacher--Student DAgger strategy that uses reliable position information. The resulting system achieves good performance in simulation.

\textbf{Limitations:} The Galaxea R1 Pro does not support PD control and uses a closed-source, high-stiffness cascaded servo controller, so our policy cannot currently be applied to the real robot. We also observe that base instability causes oscillations in the training curves and may require smoothing such as a Kalman filter. Future work will transfer the policy to another wheeled humanoid and fuse visual information to obtain more accurate localization.

\makeatletter
\let\IEEEorigthebibliography\thebibliography
\def\thebibliography#1{\IEEEorigthebibliography{#1}\fontsize{8}{8.2}\selectfont\setlength{\itemsep}{0pt}\vspace{-8pt}}
\makeatother
\bibliographystyle{IEEEtran}
\bibliography{references}

@inproceedings{h2o2024,
  author    = {Tairan He and Zhengyi Luo and Wenli Xiao and Chong Zhang and Kris Kitani and Changliu Liu and Guanya Shi},
  title     = {Learning Human-to-Humanoid Real-Time Whole-Body Teleoperation},
  booktitle = {Proc. IEEE/RSJ Int. Conf. Intell. Robots Syst. (IROS)},
  year      = {2024},
  pages     = {8944--8951},
  doi       = {10.1109/IROS58592.2024.10801984}
}

@article{omnih2o2024,
  author  = {Tairan He and Zhengyi Luo and Xialin He and Wenli Xiao and Chong Zhang and Weinan Zhang and Kris Kitani and Changliu Liu and Guanya Shi},
  title   = {{OmniH2O}: Universal and Dexterous Human-to-Humanoid Whole-Body Teleoperation and Learning},
  journal = {arXiv preprint arXiv:2406.08858},
  year    = {2024},
  doi     = {10.48550/arXiv.2406.08858}
}

@article{beyondmimic2025,
  author  = {Qiayuan Liao and Takara E. Truong and Xiaoyu Huang and Yuman Gao and Guy Tevet and Koushil Sreenath and C. Karen Liu},
  title   = {{BeyondMimic}: From Motion Tracking to Versatile Humanoid Control via Guided Diffusion},
  journal = {Science Robotics},
  volume  = {11},
  number  = {117},
  pages   = {eadx8924},
  year    = {2026},
  doi     = {10.1126/scirobotics.adx8924}
}

@article{bfmzero2025,
  author  = {Yitang Li and Zhengyi Luo and Tonghe Zhang and Cunxi Dai and Anssi Kanervisto and Andrea Tirinzoni and Haoyang Weng and Kris Kitani and Mateusz Guzek and Ahmed Touati and Alessandro Lazaric and Matteo Pirotta and Guanya Shi},
  title   = {{BFM-Zero}: A Promptable Behavioral Foundation Model for Humanoid Control Using Unsupervised Reinforcement Learning},
  journal = {arXiv preprint arXiv:2511.04131},
  year    = {2025},
  doi     = {10.48550/arXiv.2511.04131}
}

@article{du2024efficient,
  author  = {Wenqian Du and Ran Long and Jo{\~a}o Moura and Jiayi Wang and Saeid Samadi and Sethu Vijayakumar},
  title   = {Efficient Whole-Body Model Predictive Control for Online Compliant Dual-Arm Mobile Manipulation},
  journal = {Journal of Field Robotics},
  year    = {2026},
  note    = {Art. no. rob.70320},
  doi     = {10.1002/rob.70320}
}

@article{l2026generativempc,
  author  = {Marcelino Julio Fernando and Miguel Altamirano Cabrera and Jeffrin Sam and Yara Mahmoud and Konstantin N. Gubernatorov and Dzmitry Tsetserukov},
  title   = {{GenerativeMPC}: {VLM-RAG}-Guided Whole-Body {MPC} with Virtual Impedance for Bimanual Mobile Manipulation},
  journal = {arXiv preprint arXiv:2604.19522},
  year    = {2026},
  doi     = {10.48550/arXiv.2604.19522}
}

@inproceedings{qiao2025signbot,
  author    = {Guanren Qiao and Sixu Lin and Ronglai Zuo and Zhizheng Wu and Kui Jia and Guiliang Liu},
  title     = {{SignBot}: Learning Human-to-Humanoid Sign Language Interaction},
  booktitle = {Proc. IEEE Int. Conf. Robot. Autom. (ICRA)},
  year      = {2026},
  pages     = {20471--20478},
  doi       = {10.1109/ICRA57385.2026.11696513}
}

@inproceedings{wang2025seec,
  author    = {Jaehwi Jang and Zhuoheng Wang and Ziyi Zhou and Feiyang Wu and Ye Zhao},
  title     = {{SEEC}: Stable End-Effector Control with Model-Enhanced Residual Learning for Humanoid Loco-Manipulation},
  booktitle = {Proc. IEEE Int. Conf. Robot. Autom. (ICRA)},
  year      = {2026},
  pages     = {1989--1996},
  doi       = {10.1109/ICRA57385.2026.11697024}
}

@inproceedings{luo2023perpetual,
  author    = {Zhengyi Luo and Jinkun Cao and Alexander W. Winkler and Kris Kitani and Weipeng Xu},
  title     = {Perpetual Humanoid Control for Real-Time Simulated Avatars},
  booktitle = {Proc. IEEE/CVF Int. Conf. Comput. Vis. (ICCV)},
  year      = {2023},
  pages     = {10861--10870},
  doi       = {10.1109/ICCV51070.2023.01000}
}

@inproceedings{araujo2025retargeting,
  author    = {Jo{\~a}o Pedro Ara{\'u}jo and Yanjie Ze and Pei Xu and Jiajun Wu and C. Karen Liu},
  title     = {Retargeting Matters: General Motion Retargeting for Humanoid Motion Tracking},
  booktitle = {Proc. IEEE Int. Conf. Robot. Autom. (ICRA)},
  year      = {2026},
  pages     = {6097--6104},
  doi       = {10.1109/ICRA57385.2026.11697263}
}

@inproceedings{araujo2025motion,
  author    = {Shiyu Liu and Ilija Had{\v{z}}i{\'c} and Akshay Gupta and Aliasghar Arab},
  title     = {Motion Planning and Control of an Overactuated 4-Wheel Drive with Constrained Independent Steering},
  booktitle = {Proc. IEEE Int. Conf. Adv. Robot. (ICAR)},
  year      = {2025},
  pages     = {122--128},
  doi       = {10.1109/ICAR65334.2025.11338621}
}

@article{zhang2025hybrid,
  author  = {Runjiao Bao and Lin Zhang and Tianwei Niu and Haoyu Yuan and Shoukun Wang},
  title   = {Hybrid {A*} Path Planning with Multi-Modal Motion Extension for Four-Wheel Steering Mobile Robots},
  journal = {arXiv preprint arXiv:2509.06115},
  year    = {2025},
  doi     = {10.48550/arXiv.2509.06115}
}

@article{wang2026clot,
  author  = {Tengjie Zhu and Guanyu Cai and Zhaohui Yang and Guanzhu Ren and Haohui Xie and ZiRui Wang and Junsong Wu and Jingbo Wang and Xiaokang Yang and Yao Mu and Yichao Yan},
  title   = {{CLOT}: Closed-Loop Global Motion Tracking for Whole-Body Humanoid Teleoperation},
  journal = {arXiv preprint arXiv:2602.15060},
  year    = {2026},
  doi     = {10.48550/arXiv.2602.15060}
}

@inproceedings{mahmood2019amass,
  author    = {Naureen Mahmood and Nima Ghorbani and Nikolaus F. Troje and Gerard Pons-Moll and Michael J. Black},
  title     = {{AMASS}: Archive of Motion Capture As Surface Shapes},
  booktitle = {Proc. IEEE/CVF Int. Conf. Comput. Vis. (ICCV)},
  year      = {2019},
  pages     = {5441--5450},
  doi       = {10.1109/ICCV.2019.00554}
}

@inproceedings{ross2010dagger,
  author    = {St{\'e}phane Ross and Geoffrey J. Gordon and J. Andrew Bagnell},
  title     = {A Reduction of Imitation Learning and Structured Prediction to No-Regret Online Learning},
  booktitle = {Proc. 14th Int. Conf. Artif. Intell. Stat. (AISTATS)},
  volume    = {15},
  year      = {2011},
  pages     = {627--635},
}
\end{document}